%% file: main.tex
\documentclass{article}

\usepackage{microtype}
\usepackage{graphicx}
\usepackage{subfigure}
\usepackage{booktabs} 

\usepackage[unicode]{hyperref}

\usepackage[accepted]{icml_codeml_2025}

\usepackage{amsmath}
\usepackage{amssymb}
\usepackage{mathtools}
\usepackage{amsthm}
\usepackage{bm}
\usepackage{nicefrac}

\usepackage{algorithm}
\usepackage{algorithmic}

\usepackage[capitalize,noabbrev]{cleveref}
\usepackage{float}
\usepackage{caption}
\usepackage{stfloats}
\usepackage{titlesec}

\newfloat{codesnippet}{htbp}{loc}[section]
\floatname{codesnippet}{Code Snippet}
\floatstyle{plain}            
\restylefloat{codesnippet}    

\theoremstyle{plain}

\theoremstyle{definition}

\theoremstyle{remark}

\input{preamble}

\input{preamble_laplax}
\input{preambles/code}

\titlespacing*{\paragraph}{0pt}{0.05ex}{0.5ex}
\icmltitlerunning{\laplax\ -- Laplace Approximations with JAX}

\begin{document}

\twocolumn[
\icmltitle{\laplax \\
           Laplace Approximations with JAX}



\icmlsetsymbol{equal}{*}

\begin{icmlauthorlist}
\icmlauthor{Tobias Weber}{equal,yyy}
\icmlauthor{Bálint Mucsányi}{equal,yyy}
\icmlauthor{Lenard Rommel}{yyy}
\icmlauthor{Thomas Christie}{yyy}
\icmlauthor{Lars Kasüschke}{yyy}\\
\icmlauthor{Marvin Pförtner}{yyy}
\icmlauthor{Philipp Hennig}{yyy}
\end{icmlauthorlist}

\icmlaffiliation{yyy}{Tübingen AI center, University of Tübingen, Tübingen, Germany}

\icmlcorrespondingauthor{Tobias Weber}{t.weber@uni-tuebingen.de}

\icmlkeywords{Machine Learning, Bayes, Laplace Approximation, JAX, ICML}

\vskip 0.3in
]



\printAffiliationsAndNotice{\icmlEqualContribution} 
\renewcommand{\icmlEqualContribution}{\mbox{\textsuperscript{*}}}

\begin{abstract}
The Laplace approximation provides a scalable and efficient means of quantifying weight-space uncertainty in deep neural networks, enabling the application of Bayesian tools such as predictive uncertainty and model selection via Occam's razor. In this work, we introduce \laplax\footnote{\url{https://github.com/laplax-org/laplax}}, a new open-source Python package for performing Laplace approximations with \texttt{jax}. Designed with a modular and purely functional architecture and minimal external dependencies, \laplax offers a flexible and researcher-friendly framework for rapid prototyping and experimentation. Its goal is to facilitate research on Bayesian neural networks, uncertainty quantification for deep learning, and the development of improved Laplace approximation techniques.
\end{abstract}

\input{sections/introduction}

\input{sections/background}

\input{sections/toolbox}

\input{sections/experiments}

\input{sections/conclusion}

\section*{Acknowledgements}

The authors gratefully acknowledge financial support by the European Research Council through ERC CoG Action 101123955 ANUBIS; the DFG Cluster of Excellence "Machine Learning - New Perspectives for Science", EXC 2064/1, project number 390727645; the German Federal Ministry of Education and Research (BMBF) through the Tübingen AI Center (FKZ: 01IS18039A); the DFG SPP 2298 (Project HE 7114/5-1), and the Carl Zeiss Foundation (project "Certification and Foundations of Safe Machine Learning Systems in Healthcare"), as well as funds from the Ministry of Science, Research and Arts of the State of Baden-Württemberg. 

\section*{Impact Statement}

This paper describes a software library designed to facilitate research in approximate Bayesian uncertainty quantification for deep neural networks.
There are many potential societal consequences of our work, as predictive uncertainty quantification contributes to mitigating the risk of real-world AI applications.

\bibliography{references}
\bibliographystyle{icml2025}

\newpage
\appendix
\onecolumn

\input{appendix/computational_details}

\input{appendix/related_work}

\input{appendix/fsp_laplace}

\input{appendix/classification}




\end{document}

%% file: preamble.tex
\newcommand{\cC}{\mathcal{C}}
\newcommand{\cD}{\mathcal{D}}

\newcommand{\cM}{\mathcal{M}}
\newcommand{\cN}{\mathcal{N}}

\DeclarePairedDelimiterX{\set}[1]{\{}{\}}{%
  #1%
}

\DeclarePairedDelimiterXPP{\restrict}[3]{}{.}{\rvert}{_{#2}^{#3}}{#1}  

\newcommand{\R}{{\mathbb{R}}}  

\DeclarePairedDelimiterX{\interval}[2]{[}{]}{#1,#2}
\DeclarePairedDelimiterX{\ointerval}[2]{]}{[}{#1,#2}
\DeclarePairedDelimiterX{\lointerval}[2]{]}{]}{#1,#2}
\DeclarePairedDelimiterX{\rointerval}[2]{[}{[}{#1,#2}

\DeclarePairedDelimiterXPP{\pvidx}[2]{}{(}{)}{_{#2}}{#1}  

\newcommand*{\vecsym}[1]{\boldsymbol{#1}}

\newcommand{\vv}{\vecsym{v}}

\newcommand{\vmu}{\vecsym{\mu}}

\newcommand{\mat}[1]{{\boldsymbol{#1}}}

\newcommand{\mA}{\mat{A}}

\newcommand{\mD}{\mat{D}}

\newcommand{\mH}{\mat{H}}
\newcommand{\mI}{\mat{I}}
\newcommand{\mJ}{\mat{J}}

\newcommand{\mL}{\mat{L}}
\newcommand{\mM}{\mat{M}}

\newcommand{\mS}{\mat{S}}

\newcommand{\mU}{\mat{U}}
\newcommand{\mV}{\mat{V}}

\newcommand{\mLambda}{\mat{\Lambda}}

\newcommand{\mSigma}{\mat{\Sigma}}

\DeclarePairedDelimiterX{\inprod}[2]{\langle}{\rangle}{#1, #2}  

\DeclareMathOperator*{\argmin}{arg\,min}
\DeclareMathOperator*{\argmax}{arg\,max}

\newcommand{\given}{\vert}

\newcommand{\atgiven}[1]{%
  \nonscript\:#1\vert
  \nonscript\:
  \mathopen{}%
}

\DeclarePairedDelimiterX{\condps}[1]{(}{)}{%
  \renewcommand{\given}{\atgiven{\delimsize}}%
  #1%
  \renewcommand{\given}{\vert}%
}

\newcommand{\pdens}[2][\pdf]{#1\condps*{#2}}

\DeclarePairedDelimiterX{\condbrks}[1]{[}{]}{%
  \renewcommand{\given}{\atgiven{\delimsize}}%
  #1%
}



%% file: preamble_laplax.tex
\newcommand{\inputs}{\ensuremath{x}}  
\newcommand{\targets}{\ensuremath{y}} 
\newcommand{\params}{\ensuremath{\theta}} 
\newcommand{\mapparam}{\ensuremath{\theta}^*}
\newcommand{\traininputs}{\ensuremath{X}}
\newcommand{\traintargets}{\ensuremath{Y}}

\newcommand{\traindata}{\ensuremath{\mathcal{D}}} 
\newcommand{\contextpoints}{\ensuremath{\mathcal{C}}} 
\newcommand{\batch}{\ensuremath{\mathcal{B}}} 
\newcommand{\contextpoint}{\ensuremath{\mathbf{c}_i}}

\newcommand{\dimparams}{\ensuremath{P}}
\newcommand{\numtraindata}{\ensuremath{N}}

\newcommand{\network}{\ensuremath{f}} 

\newcommand{\loss}{\ensuremath{\ell}} 
\newcommand{\risk}{\ensuremath{\mathcal{L}}} 
\newcommand{\regularizer}{\ensuremath{\Omega}} 
\newcommand{\fsprisk}{\ensuremath{R}}
\newcommand{\rkhs}{\ensuremath{\mathbb{H}_{\Sigma}}}
\newcommand{\rkhsnorm}{\ensuremath{\|\cdot\|_{\rkhs}}}

\newcommand{\posteriorprecision}{\mH_{\params^*}}

\newcommand{\jacmap}{\ensuremath{\mJ_{\theta^*} }}


%% file: preambles/code.tex
\usepackage{xcolor}
\usepackage[listings]{tcolorbox}
\usepackage{listings}
\usepackage{caption}
\usepackage{xspace}

\definecolor{laplaxcolor}{HTML}{00695B}
\DeclareRobustCommand{\laplax}{%
  \textcolor{laplaxcolor}{\texttt{laplax}}\xspace%
}

\definecolor{tue_dark}{HTML}{36404A}        
\definecolor{tue_darkblue}{HTML}{405A8C}    
\definecolor{tue_blue}{HTML}{0069AA}        
\definecolor{tue_green}{HTML}{7DA44B}       
\definecolor{tue_gray}{HTML}{AFB2B7}        
\definecolor{tue_lightgold}{HTML}{EEECE2}   
\definecolor{tue_lightblue}{HTML}{4FAAC8}   
\definecolor{tue_lightgreen}{HTML}{81B99F}  

\tcbset{
  listing engine=listings,
  colback=tue_lightgold!60,   
  colframe=tue_lightgold!60,  
  coltitle=tue_darkblue,      
  fonttitle=\bfseries,
  top=1mm, bottom=1mm, left=1mm, right=1mm,
  arc=1mm,
  boxsep=1mm,
}

\lstdefinelanguage{MyPython}{
  morekeywords={False, None, True, and, as, assert, async, await, break, class, continue,
    def, del, elif, else, except, finally, for, from, global, if, import, in,
    is, lambda, nonlocal, not, or, pass, raise, return, try, while, with, yield},
  sensitive=true,
  morecomment=[l]\#,
  morestring=[b]',
  morestring=[b]",
}

\lstdefinestyle{tuebingen}{
  language=MyPython,
  backgroundcolor=\color{tue_lightgold!60},
  basicstyle=\ttfamily\small,
  keywordstyle=\color{tue_blue}\bfseries,
  commentstyle=\itshape\color{tue_darkblue},
  stringstyle=\color{tue_green},
  numberstyle=\tiny\color{tue_gray},
  numbers=none,
  numbersep=0pt,
  frame=none,
  showstringspaces=false,
  breaklines=true,
  postbreak=\raisebox{0ex}[0ex][0ex]{\(\hookrightarrow\)},
  xleftmargin=3ex,
  xrightmargin=1ex,
  captionpos=b,
  mathescape=true,
}

\newtcblisting{codeblock}[2][]{%
  listing only,
  listing options={
    style=tuebingen,
    language=MyPython,
    breaklines,
    #1
  },
  top=0pt, bottom=0pt, left=0pt, right=0pt,
  boxrule=0pt,            
  colback=tue_lightgold!60,
  colframe=tue_lightgold!60,
  before skip=0pt,
  after skip=0pt,
  arc=0mm,           
}

\newcommand{\codeinline}[1]{%
  \colorbox{tue_lightgold!60}{\ttfamily\small #1}%
}

%% file: sections/introduction.tex
\section{Introduction}
\label{sec:introduction}
\begin{figure}[ht]
    \includegraphics{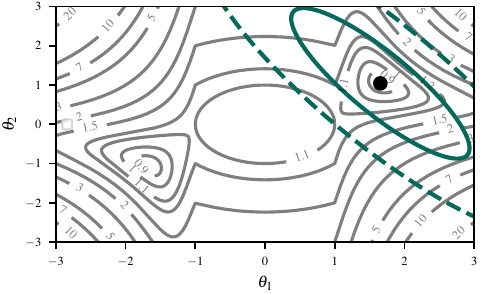}
    \caption{Linearised Laplace approximation on a two-parameter ReLU network $\network(x,\theta)=\theta_2\,\text{ReLU}(\theta_1 x+1)$ trained on $\cD=\{(1,-1),(-1,-1)\}$.
    Gray contours: energy with square loss; black dot: optimised weights $\params^*$; green ellipses: $1\sigma$ and $2\sigma$ levels of the Laplace approximation.}
    \label{fig:laplace_relu}
\end{figure}

\begin{codesnippet}[ht]

\begin{codeblock}{}
from jax.nn import relu
from jax.numpy import array
from laplax import laplace
from plotting import plot_figure_1

# You need a model...
def model_fn(input, params):
    return relu(
        params["theta1"] * input - 1
    ) * params["theta2"]
params = { # optimized weights,
    "theta1": array(1.6556547), 
    "theta2": array(1.0420421)
}
data = {  # and training data.
    "input": array([1., -1.]),
    "target": array([1., -1.])
}
# ... then apply laplax ...
posterior_fn, _ = laplace(
    model_fn, params, data,
    loss_fn="mse", curv_type="full",
)
arg = {"prior_prec": 0.2}
curv = posterior_fn(arg).state['scale']
# ... to get Figure 1.
plot_figure_1(model_fn, params, curv)
\end{codeblock}
\label{code:laplace_relu}
\caption{\laplax\ code for generating \cref{fig:laplace_relu}.}
\end{codesnippet}
Bayesian modelling provides principled approaches to several open challenges in modern deep learning \cite{papamarkou2024positionbayesiandeeplearning}, including overconfidence in predictions \cite{kristiadi_being_2020}, catastrophic forgetting in continual learning \cite{ritter_online_2018}, and the incorporation of prior knowledge into model predictions~\cite{cinquin_fsp_laplace_2024}. 
The Laplace approximation \cite{mackay1992practical} offers a computationally efficient, post-hoc method for approximating the posterior distribution over neural network weights, effectively transforming standard deep architectures into Bayesian neural networks.
This enables the use of Bayesian tools such as predictive uncertainty estimation, marginal likelihood evaluation, and model selection.

Despite its conceptual simplicity, implementing the Laplace approximation involves several non-trivial choices, ranging from curvature estimation and posterior parameterization to calibration and inference techniques. 
While a comprehensive implementation exists for PyTorch~\cite{daxberger_laplace_2021}, a similarly extensive but more flexible and research-oriented solution for the \texttt{jax} ecosystem has been lacking.

To address this gap, we introduce \laplax, a lightweight and modular Python library for Laplace approximations built entirely on \texttt{jax} \cite{jax2018github}.
Designed with research flexibility in mind, \laplax\ supports seamless integration with any \texttt{jax}-based deep learning framework. 
It features both a high-level, functional API for rapid experimentation (see Code Snippet 1 producing \cref{fig:laplace_relu}) and low-level building blocks to support in-depth analyses and changing the algorithm itself.

In this paper, we outline the design principles of \laplax, describe its core components, and demonstrate its application on a simple regression and classification task.

%% file: sections/background.tex
\section{Making Neural Networks Bayesian}
\label{sec:background}

Given labelled training data $\traindata = \{(\inputs_n, \targets_n)\}_{n=1}^\numtraindata$, loss function $\loss$ and regularizer $\regularizer$, the parameters $\params$ of a neural network $\network_\params$ are typically obtained by minimising the regularised empirical risk $\risk(\traindata, \network_\params)$. 
From a probabilistic perspective, this procedure corresponds to finding a maximum a posteriori (MAP) estimate of the weights under a likelihood and prior.
Formally, both views lead to the following optimisation problem:
\begin{align*}
\params^* &= \argmin_{\params} \mathcal{L} (\mathcal{D}, \network_\params) \\
          &= \argmin_{\params} \underbrace{\sum_{n=1}^{\numtraindata} \loss(\network(\inputs_n, \params), \targets_n) + \regularizer(\params)}_{\substack \risk(\traindata, \network_\params)} \\
          &= \argmax_{\params} \sum_{n=1}^{\numtraindata} \log \pdens{\targets_n \given \network(\inputs_n, \params)} + \log \pdens{\params}\, .
\end{align*}
The weight-space uncertainty is then described by the posterior distribution given the training data:
\begin{align*}
\pdens{\params \given \traindata} = \frac{ \pdens{\traindata \given \params}\, \pdens{\params}}{\int \pdens{\traindata \given \params}\,\pdens{\params}\,d\params}\; .
\end{align*}
However, for deep neural networks, the integral in the denominator is usually intractable. The Laplace approximation circumvents this by utilising a Gaussian distribution to approximate the posterior. 
To this end, we apply a second-order Taylor approximation to the negative log-posterior loss $\risk$ around the MAP estimate $\params^*$, which yields
\begin{align*}
\risk(\traindata, \network_\params) \approx &\risk(\traindata, \network_{\params^*}) + {\color{tue_gray}\nabla_\theta \risk(\traindata, \network_{\params^*})^\top (\params - \params^*)} \\&+ \frac{1}{2} (\params - \params^*)^\top\nabla^2_{\params \params} \risk(\traindata, \network_{\params^*}) (\params - \params^*),
\end{align*}
where the \textcolor{tue_gray}{first-order term} vanishes due to the assumed local optimality of $\params^*$. Negation and exponentiation yield
\begin{equation}
\label{eq:weightposterior}
\pdens{\params \given \traindata} \approx \cN\Bigl(\theta^*, \posteriorprecision^{-1}\Bigr)
\end{equation}
with $\posteriorprecision =  \nabla^2_{\params\params} \risk(\traindata, \network_{\params^*})$ being the posterior precision.
To obtain predictive uncertainty estimates, the weight space uncertainty is pushed forward into the neural network's output space. This is either done via sampling a set of $S$ weights from the approximate posterior and using these in the neural network forward pass to obtain $S$ predictions, or by \emph{linearising} the network around the MAP estimate as
\begin{equation*}
\network^{\text{lin}}(\cdot, \params) = \network(\cdot, \params^*) + \mathcal{J}_{\params^*}(\cdot)(\params - \params^*)
\end{equation*} 
and using the closure properties of Gaussian distributions under affine maps \citep{immer_improving_2021}, yielding \emph{closed-form} output-space uncertainty.\footnote{For classification, the logit-space uncertainty is analytic, but the predictive distribution has to be approximated, e.g., through Monte Carlo sampling and averaging the softmax probabilities.} 
The linearised approach is guaranteed to yield positive-definite weight-space covariance matrices for a strictly convex regulariser $\Omega$ at any weight configuration $\theta$, not just at MAP estimates (that are hard to obtain exactly in deep learning settings).
Usually, further approximations are needed to reduce the computational and memory requirements of the curvature. These are discussed in~\cref{par:curv_approx}.

An important Bayesian tool for model selection is the log marginal likelihood given by
\begin{equation}
\label{eq:lml}
\log \pdens{\traindata \given \cM} \approx \textcolor{tue_blue}{\log \pdens{\traindata, \params^* \given \cM}} - \textcolor{tue_green}{\frac{1}{2} \log \left\vert \frac{1}{2\pi} \mH_{\params^*} \right\vert}.
\end{equation}
This term is often used for the selection of the model hyperparameters $\cM$ via maximization \citep{immer_scalable_2021}, since it represents an analytic trade-off between \textcolor{tue_green}{complexity} and \textcolor{tue_blue}{expressivity} -- the so-called Occam's razor \cite{rasmussen2000occam}. Tractability and scalability depend on the structure of the estimated $\posteriorprecision$, but compared to the predictive uncertainty above (cf. ~\Cref{eq:weightposterior}), no inversion is needed.

%% file: sections/toolbox.tex
\section{\laplax ---A Modular Toolbox}
\label{sec:toolbox}
The high-level API of \laplax provides the main function \codeinline{laplace(\dots)} for fitting weight space curvature approximations (see Code Snippet 1 for usage), as well as additional functions for calibrating hyperparameters (\codeinline{calibration(\dots)}) and a framework for evaluations (\codeinline{evaluation(\dots)}).
The latter two operate on the weight-space posterior function \codeinline{posterior\_fn}, which maps hyperparameters, e.g., the prior precision, to the posterior covariance.
While these high-level functions enable quick experimentation, the core strength of \laplax lies in its modular design. Each step of the Laplace approximation pipeline is exposed as an independent, composable function, closely reflecting the general algorithmic scaffold. 
These components are designed to be easily replaceable, encouraging experimentation and replacement with alternative functionality from complementary packages (e.g.,~\cite{Pinder2022}).
In the following, we provide an overview of the currently available features. Complementary details are listed in the \Cref{app:computational_details}.
\paragraph{Curvature-vector products.}
A central element of the Laplace approximation is the choice of curvature. Following the motivation in \cite{dangel2025positioncurvaturematricesdemocratized}, \laplax represents and handles all curvatures as {\em matrix-vector products}. This matrix-free approach significantly reduces memory usage, enhances computational efficiency, and improves flexibility.
Currently, \laplax supports both Hessian- and Generalized Gauss-Newton (GGN)-vector products, which can be computed on arbitrary iterables of data points, including PyTorch DataLoaders~\cite{pytorch} and TensorFlow Datasets~\cite{TFDS}. 
\paragraph{\codeinline{CurvApprox}: from curvature to posterior precision.}
\label{par:curv_approx}
The primary trade-off between speed and accuracy lies in how the curvature is approximated and the structure it assumes. Once the curvature approximation method (\codeinline{CurvApprox}) is selected, it is processed into a function (\codeinline{posterior\_fn}) that returns the posterior precision ($\posteriorprecision$) \textit{matrix-vector product} given hyperparameters. These include the prior precision -- modelled as an identity matrix scaled by the \codeinline{prior\_prec} ($\tau$), representing an isotropic Gaussian prior -- and additional hyperparameters $\mathcal{C}$ of the negative log-likelihood loss, such as the observation noise $\sigma^2$ for regression~\cite{daxberger_laplace_2021}.
\laplax currently supports the following curvature approximations:
\begin{itemize}
    \item \codeinline{CurvApprox.FULL} materializes the full matrix in memory by applying the curvature-vector product to the columns of an identity matrix. The posterior function is then given by
    \begin{equation}
        (\tau, \cC) \mapsto \left[ v \mapsto \left(\textbf{Curv}(\cC) + \tau \mI \right)^{-1} v \right].
    \end{equation}
    \item \codeinline{CurvApprox.DIAGONAL} approximates the curvature using only its diagonal, obtained by evaluating the curvature-vector product with standard basis vectors from both sides. This leads to:
    \begin{equation}
        (\tau, \mathcal{C}) \mapsto \left[ v \mapsto \big(\text{diag}(\textbf{Curv}(\cC)) + \tau\mI \big)^{-1}v  \right].3
    \end{equation}
    \item \textbf{Low-Rank} employs either a custom Lanczos routine (\codeinline{CurvApprox.LANCZOS}) or a variant of the LOBPCG algorithm (\codeinline{CurvApprox.LOBPCG}). 
    These methods approximate the top eigenvectors $\mU$ and eigenvalues $\mS$ of the curvature via matrix-vector products. The posterior is then given by a low-rank plus scaled diagonal
    \begin{equation}
        (\tau, \mathcal{C}) \mapsto \left[ v \mapsto \left(\big[\mU \mS \mU^\top\big](\cC) + \tau \mI \right)^{-1} v \right].
    \end{equation}
\end{itemize}
In addition to the \codeinline{posterior\_fn}, the main \codeinline{laplace(\dots)} function also returns the curvature estimate that is relevant, e.g., for computing the log marginal likelihood. More details on the transformations are provided in \cref{app:curvatures}.

Standard approximation variants such as {\em last-layer} or {\em sub-network Laplace} -- where only a subset of model parameters is treated probabilistically~\citep{daxberger2022bayesiandeeplearningsubnetwork} -- are package-independent, since all computations in \laplax depend on a generic model signature:
\begin{equation*}
\codeinline{model\_fn}\colon (\text{input},\; \text{params}) \mapsto \text{output},
\end{equation*}
and take arbitrary PyTrees as parameters \params. This also allows for flexible integration with arbitrary \texttt{jax}-based models.
\paragraph{\codeinline{Pushforward}: from weight space to output space.}
\laplax supports two strategies for propagating uncertainty from weight space to output space: sampling-based (\codeinline{Pushforward.NONLINEAR}) and linearisation-based (\codeinline{Pushforward.LINEAR}) pushforwards. 
In classification settings, this results in uncertainty over the {\em logits}, which must be further transformed to obtain predictive uncertainty over class probabilities, either by sampling in logit space and applying the \texttt{softmax} independently, or via analytic approximations to the integral
\begin{equation}
\label{eq:softmaxintegral}
\int \operatorname{softmax}(\network_\params(\inputs_n))\;\pdens{\network_{\params}(\inputs_n) \given \traindata} \, d\params\, .
\end{equation}
A full list of supported predictive approximations is provided in \cref{app:predictives}.
\paragraph{\codeinline{calibration(\dots)}.}~\laplax supports the calibration of all posterior hyperparameters, such as $\tau$ (prior precision) or $\sigma^2$ (observation noise). There are two primary strategies for calibration: (1) maximising the log marginal likelihood (see \cref{eq:lml}) with respect to the hyperparameters; or (2) optimising for a downstream metric such as the negative log-likelihood (regression) or the expected calibration error (classification).
The library includes a basic grid search routine over $\tau$, but all functions are differentiable, allowing the use of gradient-based optimization, e.g., via \texttt{optax} \cite{deepmind2020jax}, for calibrating all hyperparameters at the same time.
While calibration with respect to a downstream metric often yields better predictive performance, marginal likelihood maximization is computationally more efficient, as it avoids matrix inversions and pushforward computations.

\paragraph{\codeinline{evaluation(\dots)}.}~To assess the predictive quality of a Laplace approximation, \laplax provides a unified evaluation interface. This combines the pushforward step with the computation of summary statistics (e.g., mean, standard deviation, covariance) of standard uncertainty quantification metrics such as negative log-likelihood, continuously ranked probability score, and others. All components are modular and can be easily extended to support custom evaluation pipelines.

%% file: sections/experiments.tex
\section{Experiments}
\label{sec:experiments}
\begin{figure*}[ht!]
\centering
\includegraphics[width=0.95\textwidth]{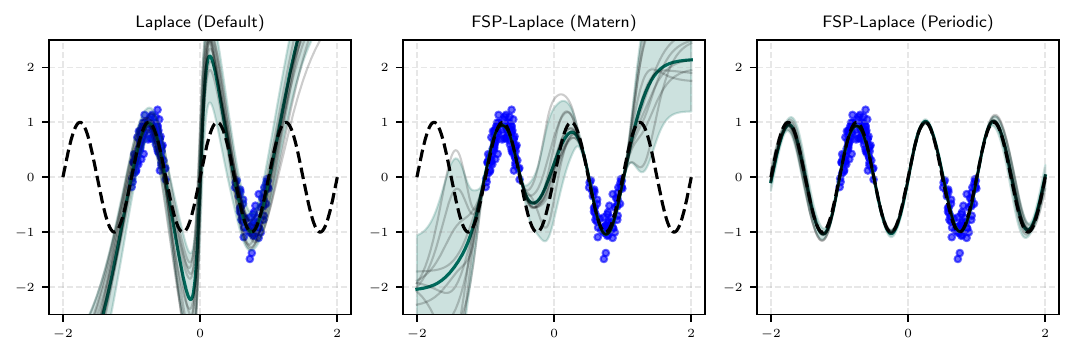}
\caption{Comparison of linearised Laplace with Lanczos-approximated curvature (NLL-GD-L) (left) and FSP-Laplace using Matérn-5/2 and periodic prior kernels. The shaded region denotes predictive uncertainty, the blue points represent training data.}
\label{fig:laplace_regression}
\end{figure*}
\input{table/table_regression}
To illustrate the practical use of \laplax, we combine its core components in a simple regression task, focusing on functionality and modularity rather than comprehensive evaluation of the quantified uncertainty. For in-depth empirical comparisons and benchmarks, refer to the Related Works section in~\Cref{app:related_work}. Here, we report only the negative log-likelihood (NLL) as a measure of calibrated uncertainty. FSP-Laplace \cite{cinquin_fsp_laplace_2024}, as a demonstration of recent advancement, is included in our visual comparison (and \laplax). All experimental code is available in the package repository; additional classification results are reported in \Cref{app:classification}.
\paragraph{Regression.}~We train a three-layer MLP with \texttt{tanh} activations and 50 hidden units on noisy samples of $y = \sin(2\pi x)$. Various curvature and calibration settings are compared by NLL in \Cref{tab:laplace_regression}. \Cref{fig:laplace_regression} visualizes predictive uncertainty for the full Laplace and FSP-Laplace (with periodic kernel and Matérn-5/2).

%% file: table/table_regression.tex
\begin{table*}
\centering
\caption{NLL results for Laplace approximation under different curvature and calibration settings (Log Marginal Likelihood objective (LML) or Negative Log-Likelihood (NLL); Grid Search (GS) or Gradient Descent (GD) -- latter also calibrating the observation noise $\sigma^2$ -- with either (Linear (L) or Nonlinear (NL) pushforward.}
\label{tab:laplace_regression}
\scalebox{0.82}{
\begin{tabular}{lcccccccc}
\toprule
 & LML-GS-L & NLL-GS-L & LML-GD-L & NLL-GD-L & LML-GS-NL & NLL-GS-NL & LML-GD-NL & NLL-GD-NL \\
\midrule
\codeinline{.FULL} (last-layer) & 1.5283 & \textbf{1.2843} & \textbf{0.4799} & \textbf{0.4482} & 1.8311 & \textbf{1.4998} & \textbf{0.5480} & \textbf{0.5400} \\
\codeinline{.FULL} & 0.8457 & 1.4409 & 0.8988 & 0.5104 & 4.2959 & 2.1243 & 4.3586 & 3.4363 \\
\codeinline{.DIAGONAL} & \textbf{0.7687} & 2.1212 & 2.2358 & 2.5784 & \textbf{1.0230} & 1.6201 & 2.5373 & 2.4986 \\
\codeinline{.LANCZOS} & 0.9437 & 1.3771 & 0.5109 & 0.5008 & 2.4008 & 2.4717 & 2.3857 & 2.8676 \\
\bottomrule
\end{tabular}
}
\end{table*}

%% file: sections/conclusion.tex
\section{Perspective and conclusion}
\label{sec:conclusion}

By having a modular, functional, and loosely-typed implementation without any hard dependency besides \texttt{jax}, we aim at a flexible framework where parts can be easily reused -- to study or implement other UQ pipelines -- or swapped with building blocks from different libraries, supporting the common research workflow. Furthermore, the framework inherits standard features from \texttt{jax}, such as \texttt{jit}-compilation, hardware-agnosticity, parallelization (\texttt{vmap}), and autodiff (\texttt{grad}, \texttt{jvp}, \texttt{vjp}).
\paragraph{Limitations and outlook.}~Some important building blocks are yet to be implemented in \laplax. These include a general Kronecker-Factored approximation for curvature-vector products (KFAC) and the family of Fisher curvature-vector products. While these curvature proxies are easy for the user to register and use (e.g., with the KFAC implementation of \cite{kfac-jax2022github}), we are currently working on providing them directly in future iterations of the library.

%% file: appendix/computational_details.tex
\section{Computational details}
\label{app:computational_details}

In the following section, we provide additional details of how different computations are performed and list more available building blocks of \laplax.

\subsection{Curvatures}
\label{app:curvatures}

The package supports both Hessian- and GGN-vector products:
\begin{align*}
v \mapsto \sum_{n=1}^{N} \nabla_{\params \params}^2 \loss(\network_\params(x_n), y_n)\,v \quad &\text{(Hessian-vector product)}, \\
v \mapsto \sum_{n=1}^{N} \mathcal{J}_\params^\top(\network_{\params^*}(x_n)) \nabla^2_{\network_\params^*(x_n),\network_\params^*(x_n)} \loss(f_{\params^*}(x_n), y_n) \mathcal{J}_\params(\network_{\params^*})\, v\quad &\text{(GGN-vector product)}
\end{align*}

We then provide the following approximation pipelines. Each of them also comes with their individual log marginal likelihood implementation.

\subsection{\codeinline{CurvApprox.FULL}}
\label{app:curv_type_full}

\paragraph{Estimation.} The total curvature $\textbf{Curv}(\cC)$ is computed by multiplying the curvature-vector product by the identity matrix.

\paragraph{Posterior precision.} The posterior precision is then computed in a dense form with
$$\posteriorprecision = \textbf{Curv}(\cC) + \tau \mI_\dimparams $$
where we $\tau \mI_\dimparams$ represent the prior with $\params \in \R^P$. 

\paragraph{Posterior scale.} We compute a left square root of the posterior covariance matrix by means of a Cholesky decomposition of the posterior precision matrix and then solve a triangular system to recover the scale matrix. This matches the implementation in \texttt{torch.distributions.multivariate.normal}~\cite{pytorch}.

\paragraph{Log marginal likelihood.} The log marginal likelihood combines the joint log-likelihood at the MAP estimate with a precision‐based evidence correction:
\begin{equation*}
\log \pdens{\traindata \given \mathcal{M}}
= \underbrace{-\frac{1}{2\sigma^2}\sum_{n=1}^N\ell(\targets_n \given \network_\params(x_n), \mathcal{M}) - \tfrac\tau2 \| \params \|^2}_{\substack{=: \pdens{\traindata, \params \given \mathcal{D}}}} -\tfrac12\Bigl(P\log(2\pi)\;-\;\log\bigl\lvert \posteriorprecision \bigr\rvert\Bigr).
\end{equation*}
Here, the joint likelihood is given by the training objective, including the prior / regularization term, including $\sigma^2$ and $\tau$, both of which would also appear in the posterior precision $\posteriorprecision$.
In regression settings, $\sigma^2$ plays the role of observation noise. It can also be viewed as a curvature hyperparameter $\cC$.

\subsubsection{\codeinline{CurvApprox.DIAGONAL}}
\label{app:curv_type_diagonal}

\paragraph{Estimation.} The diagonal curvature approximation extracts only the diagonal of the full curvature matrix $\textbf{Curv}(\cC)$, using basis-vector multiplications with the given curvature-vector product.

\paragraph{Posterior precision.} The diagonal entries of the posterior precision are formed by adding the isotropic prior precision $\tau$ to the curvature diagonal $c_i = \operatorname{diag}(\textbf{Curv}(\cC))_i$:

$$ d_i = c_i + \tau\,,\quad i=1,\dots,P.$$

\paragraph{Posterior scale.} The scale factors are simply the element-wise square‐root of the inverse precision:
\begin{equation*} 
L_{ii} \;=\;\sqrt{\frac{1}{d_i}}, 
\end{equation*}
so that $L\,L^\top$ yields the diagonal covariance.

\paragraph{Log marginal likelihood.} In the diagonal case, the evidence correction reduces to $\sum_{i=1}^P\log d_i$.  Writing the joint objective at the MAP as before, the log marginal likelihood is
\begin{equation*}
\log \pdens{\traindata \given \cM}
=
\underbrace{-\tfrac{1}{2\sigma^2}\sum_{n=1}^N\ell(\targets_n\given \network_\params(\inputs_n),\cM)
-\tfrac{\tau}{2}\|\params\|^2}_{=: \log \pdens{\traindata, \params^* \given \cM}}
-\tfrac12\Bigl(P\log(2\pi)-\sum_{i=1}^P\log d_i\Bigr).
\end{equation*}

\subsubsection{Low Rank: \codeinline{CurvApprox.LANCZOS} and \codeinline{CurvApprox.LOBPCG}}
\label{app.curv_type_low_rank}

\paragraph{Estimation.} We approximate the leading $R$ eigenpairs of the full curvature via a low‐rank method (Lanczos or LOBPCG) applied to the curvature‐vector product.  Both extract $\mU \in\R^{P\times R}$ and eigenvalues $\mS \in \R^R$, but Lanczos typically requires significantly fewer {matrix-vector products}.

\paragraph{Posterior precision.} Denoting the prior precision by $\tau$, the low‐rank posterior precision is
\begin{equation*}
    \posteriorprecision =\,\mU\,\operatorname{diag}(\mS)\,\mU^\top+\tau \mI\, .
\end{equation*}

\paragraph{Posterior scale.} We transform the posterior precision to its inverse square root by means of the procedure outlined by \citet[Section E.1]{Roy2024ReparameterizationBNN}:
\begin{equation*} 
v \mapsto \tau^{-1/2}\,v \;+\; \mU \bigl(\operatorname{diag}(\overline{\mS})\,(\mU^\top v)\bigr)\, ,
\end{equation*}
where we have
\begin{equation*}
\overline{\mS} = (\mS + \tau)^{-1/2} - \tau^{-1/2}\, .
\end{equation*} 

\paragraph{Log marginal likelihood.} The evidence correction uses the matrix determinant lemma:
$$\log\lvert H\rvert
= P\log\tau \;+\;\sum_{i=1}^R\log\bigl(1 + \tau^{-1}\mS_i\bigr),
$$
so that  
$$
\log \pdens{\traindata \given \cM}
=\log \pdens{\traindata,\params_* \given \cM}
-\tfrac12\Bigl(P\log(2\pi)-\bigl[P\log\tau+\sum_{i=1}^R\log(1+\tau^{-1}\mS_i)\bigr]\Bigr).
$$

\subsection{Pushforward and predictives}
\label{app:predictives}
We distinguish two ways of pushing forward the weight space uncertainty for a new data sample onto the output space;

\begin{itemize}
    \item \codeinline{Pushforward.LINEAR} This takes the weight space covariance $\posteriorprecision^{-1}$ and Jacobian-vector products ($\mathcal{J}_{\params}(\inputs_{new})$) to return a covariance in output space:
    \begin{align*}   
    \cN\bigg(f(\inputs_n, \params^{*}), \mathcal{J}_\params(\network(\inputs_n, \params^{*})) \posteriorprecision^{-1} \mathcal{J}_\params(\network(\inputs_n, \params^{*}))^{\top}\bigg).
    \end{align*}
    \item \codeinline{Pushforward.NONLINEAR} This samples new weights in weight space and uses the neural network with the new weight samples to get an ensemble in output space, for which then empirical estimates can be computed.
    \begin{align*}
    \network(\inputs_n, \params_s), \quad \params_s \sim \cN \big(\params_*, \posteriorprecision^{-1} \big)\, .
    \end{align*}
\end{itemize}

For classification, additional approximations are needed to push the uncertainty from logit space onto the class labels. \laplax\ supports the following predictives (\codeinline{Predictive.*}) for approximating the integral in \Cref{eq:softmaxintegral}.
\begin{itemize}
\item \codeinline{MC\_BRIDGE} Draw $z_s\sim\cN\big(\mu,\Sigma\big)$, compute $p_s=\operatorname{softmax}(z_s)$ for $s=1\ldots S$, and form $\frac1S\sum_s p_s$.
\item \codeinline{LAPLACE\_BRIDGE} Transforms the Gaussian over logits into a Dirichlet by moment‑matching (“bridge”), yielding closed‑form Dirichlet parameters and thus an analytic predictive mean. The Laplace Bridge predictive~\cite{pmlr-v180-hobbhahn22a} approximates the true predictive as follows:
\begin{equation}
    \bm{\hat p} := \frac{\frac{1}{\bm {\tilde\sigma}^2}\left(1-\frac{2}{C}+\frac{e^{\bm{\tilde\mu}}}{C^2}\sum_{c=1}^Ce^{-\tilde\mu_c}\right)}{\sum_{c=1}^C\frac{1}{\tilde\sigma^2_c}\left(1-\frac{2}{C}+\frac{e^{\bm{\tilde\mu}}}{C^2}\sum_{c'=1}^Ce^{-\tilde\mu_{c'}}\right)}
\end{equation}
where
\begin{equation}
    \bm{\tilde\mu}^2 := \sqrt{\frac{\sqrt{C/2}} {\sum_{c=1}^C\sigma^2_c}}\bm\mu,\; \bm{\tilde\sigma^2} := \frac{\sqrt{C/2}} {\sum_{c=1}^C\sigma^2_c}\bm\sigma^2.
\end{equation}
\item \codeinline{MEAN\_FIELD\_0\_PREDICTIVE}.  A zeroth‐order mean‐field (probit‐style) approximation~\cite{lu2021meanfieldapproximationgaussiansoftmaxintegral}.
\begin{align*}
   \mathbb{E}[\operatorname{softmax}_i(\bm{z})]\approx\operatorname{softmax}_i\left(\frac{\bm{\mu}}{\sqrt{1+\lambda_0\,\operatorname{diag}(\mSigma)}}\right)
\end{align*}
which rescales each mean logit by its variance.
\item \codeinline{MEAN\_FIELD\_1\_PREDICTIVE} A first‐order pairwise approximation: for each $i$ we approximate
$\Pr(z_i>z_j)$ under the bivariate Gaussian of $(z_i,z_j)$ and then normalize:
\begin{align}
\mathbb{E}[\operatorname{softmax}_i(\bm{z})]\approx \frac{1}{1 + \sum_{i \neq k} \exp \left( -\frac{(\mu_k - \mu_i)}{\sqrt{1 + \lambda_0 (\Sigma_{kk} + \Sigma_{ii})}} \right)}.
\end{align}
\item \codeinline{MEAN\_FIELD\_2\_PREDICTIVE}.  A second‐order correction that incorporates full covariance: uses all bivariate variances and covariances in the exponentiated difference integrals:
\begin{align}
\mathbb{E}[\operatorname{softmax}_i(\bm{z})]\approx \frac{1}{1 + \sum_{i \neq k} \exp \left( -\frac{(\mu_k - \mu_i)}{\sqrt{1 + \lambda_0 (\Sigma_{kk} + \Sigma_{ii} - 2\Sigma_{ik})}} \right)}.
\end{align}
\end{itemize}

Each method trades off cost versus fidelity: sampling is asymptotically exact but can be slow; the mean‑field approximations incur only $O(C^2)$ or $O(C)$ work; and the Laplace bridge often gives the best calibrated probabilities when variances are large.

%% file: appendix/related_work.tex
\section{Applications and extensions of the Laplace approximation}
\label{app:related_work}

\Cref{sec:experiments} discusses Laplace approximation with the goal of calibrated predictive uncertainty. Here, a variety of follow-up work with Laplace exists \citep{kristiadi_learnable_2021,eschenhagen_mix_la_2021,cinquin_fsp_laplace_2024} and \citet{magnani2025linearizationturnsneuraloperators} lift the method to the setting of operator learning. Notable work shows that adding some weight space uncertainty fixes overconfidence in classification networks \citep{kristiadi_being_2020}. In comparison with other uncertainty quantification methods Laplace performs comparable on the uncertainty disentanglement benchmark of 
\citep{mucsanyi_benchmarking_2025}.

In our opinion, a huge potential of linearised Laplace approximation is given by its analytic uncertainty due to the Gaussian structure. A wide field of applications opens up via the marginal log-likelihood, which provides scalable and tractable formulation of Occam's razor \cite{immer_scalable_2021}, which was successfully used for weight pruning \citep{dhahri_shaving_2024,van2023llm} or model selection, e.g. for learning layerwise equivariance \citep{ouderaa_2023_learning}. 

The analytic uncertainty structure provided by Laplace has also been used to apply filtering techniques to neural network learning with the goal of online/continual learning \cite{ritter_online_2018,sliwa_efficient_2024} or to explore Bayesian optimization \citep{kristiadi_promises_2023}.

Various other applications exist and this non-extensive list aimed only at provided some pointers for potential use cases.

%% file: appendix/fsp_laplace.tex
\section{FSP-Laplace}
\label{sec:fsp_laplace}
The BNN literature offers only a modicum of methods for eliciting informative priors, with few exceptions \citep{tran2022needgoodfunctionalprior,sun2019functionalvariationalbayesianneural}.
The commonly used isotropic Gaussian prior imposes problematic assumptions, including unimodality and weight independence that oppose the reality in neural networks and compromise uncertainty calibration and prediction reliability. As network weights are not interpretable, formulating a good prior on them is virtually impossible \citep{fortuin2022priorsbayesiandeeplearning}.
As a remedy, \emph{FSP-Laplace} \citep{cinquin_fsp_laplace_2024} extends the linearised Laplace approximation by placing interpretable Gaussian Process (GP) priors directly in function space, thereby overcoming the non-interpretability of the weight-space prior.
This yields a more refined MAP estimate and well-calibrated epistemic uncertainties when prior knowledge is available. This is particularly useful in scientific machine learning, where prior knowledge and ideas about boundary conditions and the domain are often abundant.
We therefore offer FSP-Laplace to become part of the plethora of functionalities of \laplax as a configurable inference method, allowing users to seamlessly specify
interpretable Gaussian process priors in function space and apply the Laplace approximation to their neural network models.
As can be seen in the example experiment in Figure \ref{fig:laplace_regression}, we reproduce one of the experiments of \citet{cinquin_fsp_laplace_2024}, which shows the incorporation of a periodic GP prior in function space, yielding better predictive results.

\subsection*{FSP-Laplace approximation}
FSP-Laplace consists of two major components: first, training the neural network with an RKHS regulariser to obtain the MAP estimate, described in Algorithm \ref{alg:algorithm1}, followed by the linearised Laplace approximation around this estimate, described in Algorithm \ref{alg:algorithm2}.

\subsection{FSP Training}
FSP-Laplace differs both from vanilla Laplace approximation \cite{mackay1992practical} and standard linearised Laplace approximation \cite{immer_improving_2021} in that it requires training the model with a Reproducing Kernel Hilbert Space (RKHS) $\rkhs$ regulariser. This regulariser (eq. \eqref{eq:rkhsreg}) is added to the negative log-likelihood (eq. \eqref{eq:nll}) to form the FSP objective function (eq. \eqref{eq:rfsp}). The resulting MAP estimate incorporates the functional constraints imposed by the RKHS $\rkhs$ regulariser, which is essential for the subsequent linearised approximation step.

\begin{align}
    \fsprisk_{FSP}^{(1)}(\params^{(i)})= -\frac{n}{b}\sum_{j=1}^b \log p\bigl(\targets^{(j)} \mid \network(\inputs^{(j)}, \params^{(i)})\bigr)
    \label{eq:nll}
\end{align}
The NLL is multiplied by the factor $\frac{n}{b}$ to account for the difference between the full dataset size $n$ and the mini-batch size $b$, ensuring that the regularisation strength remains consistent regardless of batch size during stochastic optimisation.
\begin{align}
    \fsprisk_{FSP}^{(2)}(\params^{(i)})= \frac{1}{2}\bigl(\network(\contextpoints^{(i)}, \params^{(i)}) - \vmu(\contextpoints^{(i)})\bigr)^\top\, \mSigma(\contextpoints^{(i)}, \contextpoints^{(i)})^{-1}\,\bigl(\network(\contextpoints^{(i)}, \params^{(i)}) - \vmu(\contextpoints^{(i)})\bigr)
    \label{eq:rkhsreg}
\end{align}
\begin{align}
     \fsprisk_{FSP}(\params^{(i)}) =  \fsprisk_{FSP}^{(1)}(\params^{(i)}) + \fsprisk_{FSP}^{(2)}(\params^{(i)}).
    \label{eq:rfsp}
\end{align}
Algorithm \ref{alg:algorithm1} implements the FSP training procedure by computing two loss components at each iteration: the NLL $\fsprisk^{(1)}_{FSP}$ on the current mini-batch, and the RKHS regularisation term $\fsprisk^{(2)}_{FSP}$ which approximates the RKHS norm $\|\cdot\|_{\rkhs}$ of the difference between the network's prediction and the prior mean at the context points. The optimiser then updates the parameters using the combined objective $\fsprisk_{FSP}^{(1)}+\fsprisk_{FSP}^{(2)}$. The selection and sampling strategy for context points $\contextpoints^{(i)}$ is discussed in Section \ref{subsec:contextpoints}.

\begin{algorithm}[!ht]
\caption{RKHS-regularised model training \citep[Algorithm 1]{cinquin_fsp_laplace_2024}}
\label{alg:algorithm1}
\begin{algorithmic}[1]
\STATE \textbf{function} \textsc{FSPLaplaceTrain}($\network$, $\params^{(0)}$, $\mathcal{GP}(\vmu, \mSigma)$, $ P_\contextpoints$, $\traindata$, $b$)
\STATE \quad $i \leftarrow 0$
\STATE \quad \textbf{for all } minibatch $\batch = (\traininputs_\batch, \traintargets_\batch)\sim \mathbb D$ of size $ b$ \textbf{do}
\STATE \quad\quad $R^{(1)}_{\mathrm{FSP}}(\params^{(i)}) \leftarrow -\frac{n}{b}\sum_{j=1}^{b} \log p(\targets_\batch^{(j)} \mid \network(\inputs_\batch^{(j)}, \params^{(i)}))$
\STATE \quad\quad Sample context points $\contextpoints^{(i)} = \{\contextpoints_j^{(i)}\}_{j=1}^{n_\contextpoints} \overset{\mathrm{i.i.d.}}{\sim}  P_\contextpoints$
\STATE \quad\quad $R^{(2)}_{\mathrm{FSP}}(\params^{(i)}) \leftarrow \frac{1}{2}(\network(\contextpoints^{(i)}, \params^{(i)}) - \vmu(\contextpoints^{(i)}))^\top \mSigma(\contextpoints^{(i)}, \contextpoints^{(i)})^{-1} (\network(\contextpoints^{(i)}, \params^{(i)}) - \vmu(\contextpoints^{(i)}))$
\STATE \quad\quad $\params^{(i+1)} \leftarrow \mathrm{optimiserStep}(\fsprisk^{(1)}_{\mathrm{FSP}} + \fsprisk^{(2)}_{\mathrm{FSP}}, \params^{(i)})$
\STATE \quad \quad $i \leftarrow i + 1$
\STATE \quad \textbf{return } $\params^{(\text{final})}$
\end{algorithmic}
\end{algorithm}

\subsection{FSP curvature estimation}
Once the MAP estimate $\params^*$ is obtained through FSP training, we compute the linearised Laplace approximation around this point:
\begin{align}
    \pdens{\params \given \traindata} \approx \mathcal{N}(\params^*, \mLambda^{-1}),
\end{align}
with
\begin{align}
    \mLambda & = \mSigma^\dagger_{\params^*} + \sum_{i=1}^n\jacmap(\inputs^{(i)})^\top\, \mL^{(i)}_{\params^*}\,\jacmap(\inputs^{(i)}),
    \quad \text{and}
    \label{eq:fspprecision} \\
    \mL_{\params^*}^{(i)} & = \nabla^2_f[-\log p(\targets^{(i)} \mid \network)]_{\network=\network(\inputs^{(i)}, \params^*)}.
    \label{eq:ggnblock}
\end{align}
Unlike the training phase, computing the Laplace approximation requires using a large number of context points to capture the prior beliefs accurately. While the prior precision $\mSigma_{\params^*}^{\dagger}$ only needs to be computed once after training, directly forming or storing $\Lambda$ becomes computationally infeasible for large networks and extensive context sets. Since the RKHS inner products in $\mSigma_{\mapparam}^{\dagger}$ do not admit closed-form expressions, we approximate the posterior covariance as $\Sigma_{\mapparam}^{\dagger} \approx \jacmap(\contextpoints)^\top \mSigma^{-1}\jacmap(\contextpoints)$. The choice of context points $\contextpoints$ for this phase is detailed in Section \ref{subsec:contextpoints}.
To address the computational challenges, Algorithm \ref{alg:algorithm2} employs an efficient procedure that avoids the explicit formation of $\Lambda$ in high dimensions.
\begin{algorithm}[t]
\caption{linearised Laplace Approximation with GP Priors \citep[Algorithm 2]{cinquin_fsp_laplace_2024}}
\label{alg:algorithm2}
\begin{algorithmic}[1]
\STATE \textbf{function} \textsc{FSP-Laplace}($\network$, $\mathcal{GP}(\vmu, \mSigma)$, $\contextpoints$, $\mathbb{D}$, $\params^{*}$)
\STATE \quad $\vv \leftarrow \jacmap(\contextpoints)\mathbf{1} / \|\jacmap(\contextpoints)\mathbf{1}\|_2$
\STATE \quad $\mL \leftarrow \mathrm{Lanczos}(\mSigma(\contextpoints, \contextpoints), \vv)$
\STATE \quad $\mM \leftarrow \jacmap(\contextpoints)^\top \mL$
\STATE \quad $(\mU_M, \mD_M, \cdot) \leftarrow \mathrm{svd}(\mM)$
\STATE \quad $\mA \leftarrow \mD_M^2 - \sum_{i=1}^{n} \mU_M^\top \jacmap(\inputs^{(i)})^\top \mL^{(i)}_{\params^\star} \jacmap(\inputs^{(i)}) \mU_M$
\STATE \quad $(\mU_A, \mD_A) \leftarrow \mathrm{eig}(\mA)$
\STATE \quad $\mS \leftarrow \mU_M \mU_A \mD_A^{-1/2}$
\STATE \quad Find smallest $k$ s.t. $\mathrm{diag}(\jacmap(\contextpoints_i) \mS_{:,k:r} \mS_{:,k:r}^\top \jacmap(\contextpoints_i)^\top) \leq \mathrm{diag}(\mSigma(\contextpoints_i, \contextpoints_i))$ $\forall i$
\STATE \quad \textbf{return } $\mathcal{N}(\params^\star, \mS_{:,k:r} \mS_{:,k:r}^\top)$
\end{algorithmic}
\end{algorithm}
The procedure begins by precomputing the prior precision once using a Lanczos process on 
$\mSigma(\contextpoints, \contextpoints)$ to obtain $\mL$ such that $\mL\mL^\top\approx \mSigma(\contextpoints, \contextpoints)^{-1}$, avoiding repeated kernel inversions.
Then the cross-covariance $\mM=\jacmap(\contextpoints)^\top\mL$ is computed using backward-mode automatic differentiation. 
On this, we operate a singular value decomposition
$\mM = \mU_M\mD_M\mV_M^\top$
and form the matrix
$\mA =\mD_M^2 - \sum_{i=1}^n \mU_M^\top \jacmap(\inputs^{(i)})^\top L_{\mapparam}^{(i)}\jacmap(\inputs^{(i)}) \mU_M,$
where $\mL_{\mapparam}^{(i)}$ is defined in eq. \eqref{eq:ggnblock}, followed by its eigendecomposition $(\mU_A, \mD_A)$.
The untruncated posterior factor is then 
$\mS = \mU_M \mU_A \mD_A^{-1/2}$
To prevent exploding predictive variance caused by numerical instability when pseudo-inverting small eigenvalues, rank truncation is applied to regularise the posterior covariance. 
This leverages the property that, in linear-Gaussian models, the posterior precision
$\mLambda_\text{posterior} = \mLambda_\mathrm{prior} + \sum_{i=1}^n \mJ_i^\top \mL_i \mJ_i \succeq \mLambda_{prior}$ since $\mL_i \succeq 0$, implying posterior variance $\leq$ prior variance. 

Therefore, the smallest rank $k$ is selected such that the posterior marginal variance at each context point remains bounded by the prior marginal variance.
Specifically, $k$ is chosen such that for all context points 
$\text{diag}\left(\jacmap(\contextpoint) \mS_{:, k:r}\mS_{:, k:r}^\top \jacmap(\contextpoint)^\top\right) \leq \text{diag}\left(\mSigma(\contextpoint, \contextpoint)\right).$  
This constraint provides a principled approach to eliminate unstable eigenpairs, i.e., eigenvalues with small magnitudes, while preserving meaningful uncertainty quantification, ensuring the posterior remains statistically consistent with the theoretical bounds of linear-Gaussian inference.

\subsection{Choice of context points}
\label{subsec:contextpoints}
The selection and evaluation of context points is crucial for both FSP training and posterior approximation, but serves different purposes in each phase.

\textbf{Training phase:} During FSP training, context points must be kept small ($n_\contextpoints \ll n$) for computational efficiency, as the regulariser requires solving a linear system in $n_\contextpoints$ dimensions at each optimiser step. An effective strategy is to sample context points i.i.d. from a distribution $P_\contextpoints$ at every training iteration. This stochastic sampling exposes the network to diverse regularisation constraints throughout training while avoiding the computational burden of large, fixed context sets. Common choices for $P_\contextpoints$ include uniform sampling over the input domain or sampling from additional unlabeled datasets. In some cases, it even makes sense to use the training batch $\batch$ itself to compute the RKHS-regulariser $\rkhsnorm$.

\textbf{Posterior approximation:} Unlike training, computing the Laplace approximation requires a large number of context points to accurately capture prior beliefs. Since this computation occurs only once after training, computational constraints are less stringent. The context points should adequately cover the regions where predictions will be made, ensuring well-calibrated uncertainty estimates beyond the immediate vicinity of training data. Low-discrepancy sequences (e.g., Halton sequences) are often preferred for their effective coverage of high-dimensional spaces.

The key principle is that context points should span the inference domain of interest, as the function-space prior only regularises the network at these locations. Insufficient coverage can lead to poorly calibrated uncertainties in uncovered regions.

%% file: appendix/classification.tex
\section{Experiment: Classification}
\label{app:classification}

\input{table/table_classification}

For the classification experiment, we train a three-layer Convolutional Neural Network on the CIFAR-10 dataset. Similarly to the regression case, we compare different curvature approximations and calibration strategies, evaluating their performance in terms of expected calibration error (ECE).
Results are reported in \Cref{tab:classification}.

%% file: table/table_classification.tex
\begin{table}[ht]
\centering
\caption{ECE results for Laplace approximation under different curvature approximations calibrated targeting the ECE on a validation set, either with grid search (GS) or gradient descent (GD). The MAP model has an ECE of 0.0755.}
\begin{tabular}{lcc}
\toprule
\codeinline{CurvApprox} & GS & GD \\
\midrule
\codeinline{.FULL} (last layer) & 0.0762 & 0.0755 \\
\codeinline{.LANCZOS} & \textbf{0.0166} & 0.0754 \\
\codeinline{.DIAGONAL} & 0.0281 & 0.0754 \\
\bottomrule
\end{tabular}\label{tab:classification}
\end{table}